\documentclass[letterpaper, 10 pt, conference]{ieeeconf}  
\IEEEoverridecommandlockouts  

\usepackage{microtype}
\usepackage{amsmath,amssymb,amsfonts}
\usepackage{algorithmic}
\usepackage{graphicx}
\usepackage{textcomp}
\usepackage{xcolor}
\usepackage{url}
\usepackage[style=numeric, sorting=none, backref=true]{biblatex}
\DefineBibliographyStrings{english}{
  backrefpage={},
  backrefpages={}
}

\usepackage{xpatch}
\DeclareFieldFormat{backrefparens}{\addperiod{#1}}
\xpatchbibmacro{pageref}{parens}{backrefparens}{}{}
\usepackage{hyperref}
\usepackage{booktabs}
\usepackage{multirow}
\usepackage{cuted}
\usepackage{caption}
\hypersetup{
    colorlinks=true,
    linkcolor=red,
    citecolor=blue,
    filecolor=black,
    urlcolor=black,
    breaklinks=true,
    pdfborder={0 0 0}
}
\usepackage{arydshln}
\usepackage{color, colortbl}
\usepackage{float}
\def\BibTeX{{\rm B\kern-.05em{\sc i\kern-.025em b}\kern-.08em
    T\kern-.1667em\lower.7ex\hbox{E}\kern-.125emX}}

\definecolor{LightCyan}{rgb}{0.88,1,1}

\begin{document}

\title{\LARGE \bf One-Shot Dual-Arm Imitation Learning}
\author{Yilong Wang$^{1}$ and Edward Johns$^{1}$\thanks{$^{1}$ The Robot Learning Lab at Imperial College London
}}

\maketitle

\begin{abstract}
We introduce One-Shot Dual-Arm Imitation Learning (ODIL), which enables dual-arm robots to learn precise and coordinated everyday tasks from just a single demonstration of the task. ODIL uses a new three-stage visual servoing (3-VS) method for precise alignment between the end-effector and target object, after which replay of the demonstration trajectory is sufficient to perform the task. This is achieved without requiring prior task or object knowledge, or additional data collection and training following the single demonstration. Furthermore, we propose a new dual-arm coordination paradigm for learning dual-arm tasks from a single demonstration. ODIL was tested on a real-world dual-arm robot, demonstrating state-of-the-art performance across six precise and coordinated tasks in both 4-DoF and 6-DoF settings, and showing robustness in the presence of distractor objects and partial occlusions. Videos are available at: \begingroup
\hypersetup{urlcolor=blue}
\url{https://www.robot-learning.uk/one-shot-dual-arm}.
\endgroup
\end{abstract}


\section{Introduction}
Efficient dual-arm robot manipulation remains very challenging in robotics, particularly for tasks that demand precise and coordinated actions. Recent advancements in imitation learning \cite{zhao2023learning, chi2024diffusionpolicy} have significantly improved the acquisition of visuomotor skills through human demonstrations. However, these methods are often data-intensive, due to the high dimensionality of the state-action space. For dual-arm manipulation, tasks typically require fifty \cite{zhao2023learning} to thousands \cite{kim2021transformer} of demonstrations per task. Furthermore, learning dual-arm coordination from human demonstrations remains an open problem due to the complex spatial and temporal correlations \cite{liu2023birp}. This leads to our first research question: \textbf{(1) Can a dual-arm robot learn precise and coordinated tasks from a single demonstration?}

To achieve one-shot efficiency, imitation learning can be modeled as an alignment between the end-effector (EE) and target object, followed by trajectory transfer or replay of the demonstration trajectory \cite{johns2021coarse, vitiello2023one, papagiannis2024miles}. Within this paradigm, the main challenge now is to extract a coordinated dual-arm trajectory from a single demonstration, and to perform precise alignment without the need for extensive real-world data collection following the demonstration. Alignment can be achieved using visual servoing to align the EE and the target object in the same way as observed during the demonstration. To avoid the need for prior object knowledge and CAD models for pose estimation, \cite{johns2021coarse, papagiannis2024miles} performed visual servoing using self-supervised learning. However, such methods require significant additional data collection beyond the demonstration. This leads to two further research questions: \textbf{(2) How can a coordinated trajectory be derived from a single demonstration? (3) How can precise and robust visual servoing be achieved without relying on object models or additional data collection?}

\begin{figure}[t]
    \setlength{\belowcaptionskip}{-15pt}
    \centering
    \includegraphics[width=0.485\textwidth]{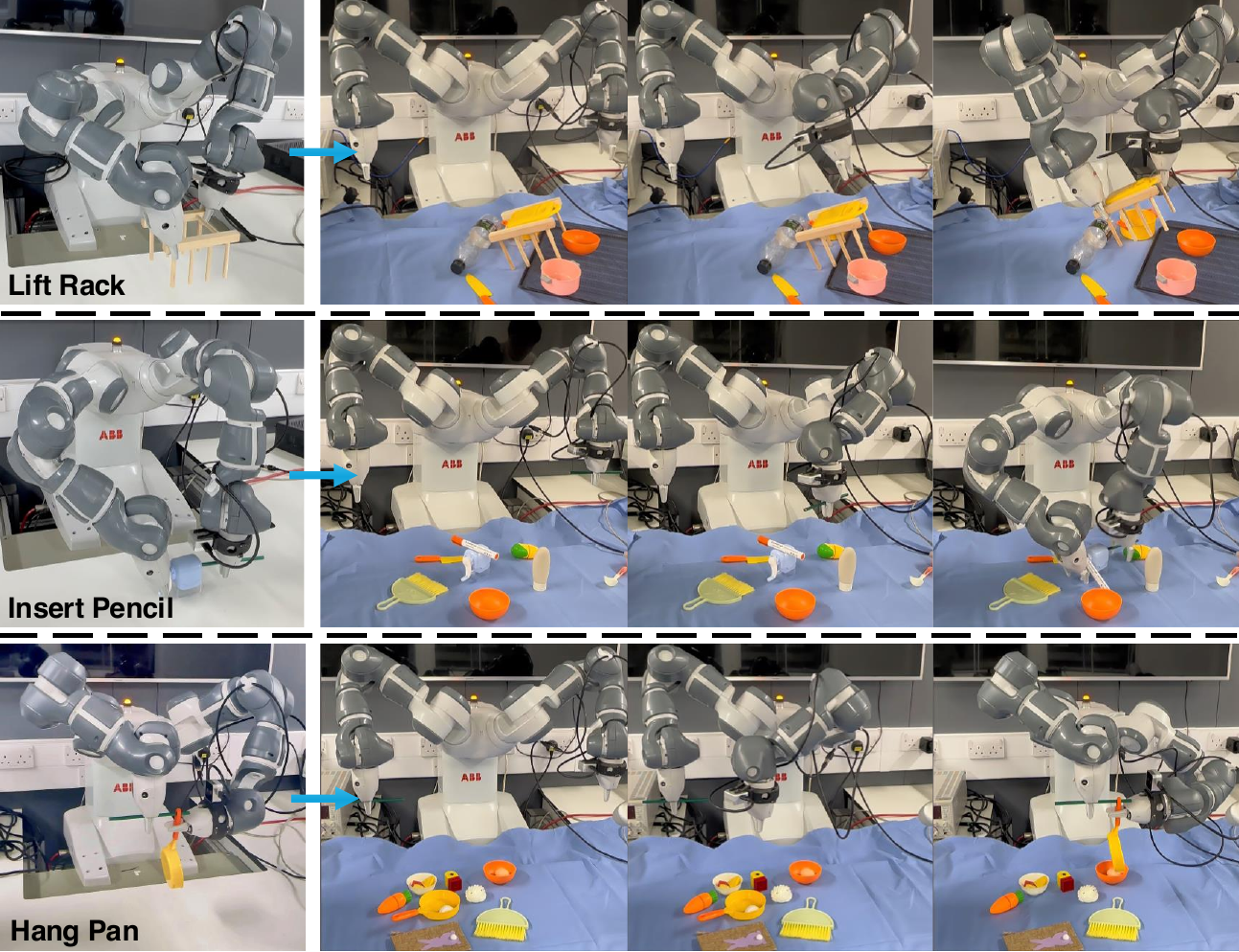}
\caption{ODIL learns precise, coordinated tasks from a single demonstration (first column) and adapts to novel robot, object, and scene configurations. Starting from an arbitrary position (second column), the robot first aligns with the object (third column), and then replays the demonstration (last column).}
    \label{fig:teaser}
\end{figure}

\begin{figure*}[t]
\setlength{\belowcaptionskip}{-15pt}
\centering
\includegraphics[width=.9\textwidth]{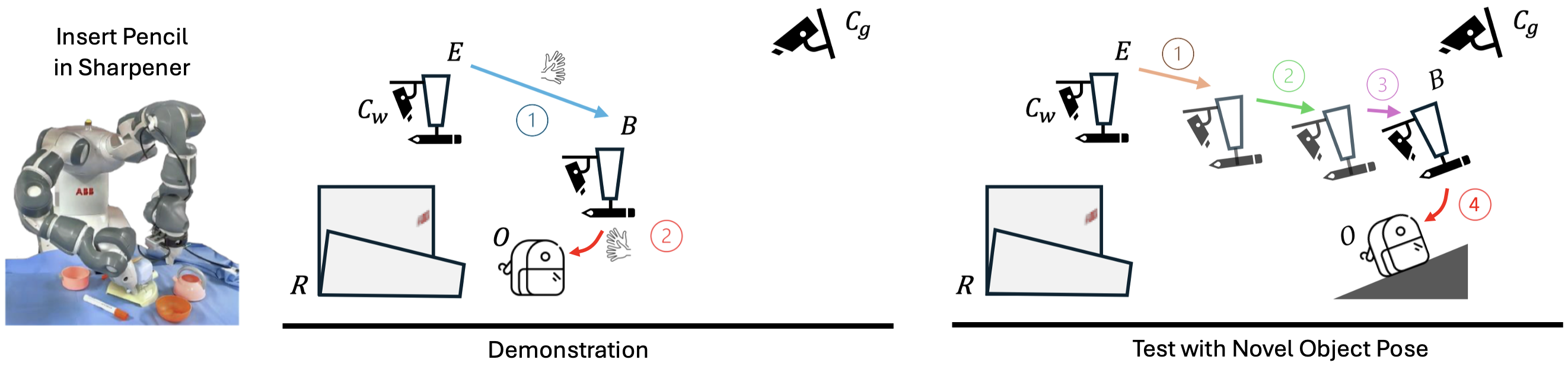}
\caption{\textbf{Overall Framework.} During demonstration, the end-effector \(E\) is first moved to a bottleneck \( {B} \), with the task object \( O \) visible to both the global and wrist cameras, \( {C}_{g} \) and \( {C}_{w} \), respectively. Then, from each camera, a bottleneck RGB-D image is captured and segmented, and a demonstration trajectory is recorded and parameterized into a coordinated trajectory \(\boldsymbol{\tau}\). During testing, starting from an arbitrary initial robot pose where \( O \) is visible to \( {C}_{g} \), we use our 3-VS controller to align \(E\) with the new \( {B} \), then execute \(\boldsymbol{\tau}\) to complete the task.}
\label{overall}
\end{figure*}


Through answering questions (1) to (3), we introduce \textbf{O}ne-shot \textbf{D}ual-arm \textbf{I}mitation \textbf{L}earning (ODIL), a method that enables a dual-arm robot with a basic position controller to learn precise, coordinated tasks from a single demonstration (Figure \ref{fig:teaser}). At its core, ODIL features a novel \textbf{Dual-Arm Coordination Paradigm}, which parametrizes a coordinated object interaction trajectory from the demonstration, and a \textbf{Three-Stage Visual Servoing (3-VS)} controller for highly precise and robust visual alignment. Once servoing is complete, the parameterized dual-arm trajectory is adapted and executed. 3-VS relies solely on one eye-to-hand global camera and one eye-in-hand wrist camera, both with only approximate hand-eye calibration. It is inspired by recent advances in deep feature matching \cite{lindenberger2023lightglue, potje2024xfeat, wang2024efficient}. Unlike modern VS methods requiring object-specific pre-training \cite{johns2021coarse} or simulation-based training \cite{valassakis2022dome}, and those that do not model outliers \cite{argus2020flowcontrol, dipalo2024dinobot, vecerik2024robotap}, we revisit the fundamentals of analytic VS \cite{wilson1996relative, Malis200021D, 999646, chaumette2007visual} and show that combining this with task-agnostic deep matchers leads to the best performance.


In our experiments, we rigorously benchmarked recent deep feature matching methods from the computer vision community for visual servoing. We then compared our imitation learning method to state-of-the-art methods \cite{dipalo2024dinobot, vecerik2024robotap}, achieving significantly better performance in 4-DoF and 6-DoF real-world tasks across diverse scenes.




\section{Related Work}

\textbf{Dual-arm Imitation Learning} involves a high-dimensional action space, which is often managed by parameterized movement primitives \cite{chitnis2020efficient, xie2020deep}. However, these primitives rely on labor-intensive, hand-crafted motions, making them difficult to scale for everyday tasks \cite{10341934}. In contrast, recent advances in dual-arm teleoperation \cite{zhao2023learning, wu2023gello, qin2023anyteleop, ding2024bunnyvisionprorealtimebimanualdexterous} have enabled the collection of high-quality demonstrations suitable for end-to-end training. This progress has allowed researchers to extend dual-arm imitation learning to impressive fine-manipulation tasks such as slotting a battery \cite{zhao2023learning}, cooking shrimp \cite{fu2024mobile}, and preparing tea \cite{wang2024dexcap}. Nevertheless, these methods require numerous human demonstrations, along with specialized hardware for collecting them, whereas our method needs only a single demonstration with minimal resources.

\textbf{One-Shot Imitation Learning} addresses the data scarcity problem by attempting to learn tasks from a single demonstration. \cite{finn2017one, duan2017one} propose meta-learning for one-shot imitation but need pre-training on similar tasks. Instead, more relevant to our work, \cite{johns2021coarse, papagiannis2024miles} employ self-supervised learning to train a VS to reach a bottleneck state, followed by executing the demonstration trajectory. Such a VS can be trained using CNNs \cite{johns2021coarse} or Siamese CNNs \cite{yu2019siamese}. However, these methods require extensive real-world data collection per object instance. \cite{puang2020kovis, valassakis2022dome} address this issue by training the VS entirely in simulation, but suffer from the sim-to-real gap. \cite{argus2020flowcontrol} uses learning-based optical flow for imitation learning. \cite{vitiello2023one} trains an unseen object pose estimator, \cite{biza23oneshot} trains a shape warper, and \cite{vosylius2023few, vosylius2024instant} train a graph neural network for in-context learning, but these operate only on noisy point clouds and so cannot perform precise tasks. More recently, keypoint representations from pre-trained visual foundation models, such as DINO-ViT \cite{caron2021emerging} and TAPIR \cite{doersch2023tapir}, have been applied to VS, eliminating the need for task-specific training and achieving state-of-the-art performance in one-shot and few-shot manipulation across various everyday tasks \cite{dipalo2024dinobot, dipalo2024kat, vecerik2024robotap}. However, variability and outliers among keypoints can overwhelm the robot controllers, causing oscillations and inaccuracies. In contrast, our method integrates pre-trained deep matchers \cite{lindenberger2023lightglue}, designed to reject outliers, with robust VS techniques \cite{Malis200021D, 999646}, resulting in highly precise alignment. Additionally, we utilize sensor fusion techniques with a global camera, providing flexibility in the robot's starting positions.


\section{Method}

Figure \ref{overall} presents an overview of ODIL's framework. The system comprises a dual-arm robot with a basic position controller operating within the robot frame \( R \), an eye-to-hand global camera \( C_{g} \), and an eye-in-hand wrist camera \( C_{w} \). The robot is initially manually moved to a bottleneck \( B \), which is conceptually attached to the task object \( O \) and selected to ensure \( O \) remains visible to both cameras. Then, from each camera a RGB-D bottleneck image is captured and segmented. A single demonstration is then conducted through kinesthetic teaching, resulting in a joint trajectory \( \left\{ \boldsymbol{q}_{i} \right\}_{i=0}^{n} \). Subsequently, this trajectory is processed to derive a coordinated object interaction trajectory \( \boldsymbol{\tau}(B, t) \), parameterized with respect to the bottleneck and time, in accordance with our Dual-arm Coordination Paradigm (Section \ref{bimanual_coordination}). During the testing phase, when encountering new robot, object, and environment configurations, our 3-VS controller (Section \ref{hdmvs}) uses the bottleneck images collected with \( C_{g} \) and \( C_{w} \) to reposition the robot to \( B \), thereby adapting \( \boldsymbol{\tau}(B, t) \) to the new context.

\subsection{Dual-arm Coordination Paradigm}
\label{bimanual_coordination}

\begin{figure}
    \centering
    \includegraphics[width=0.41\textwidth]{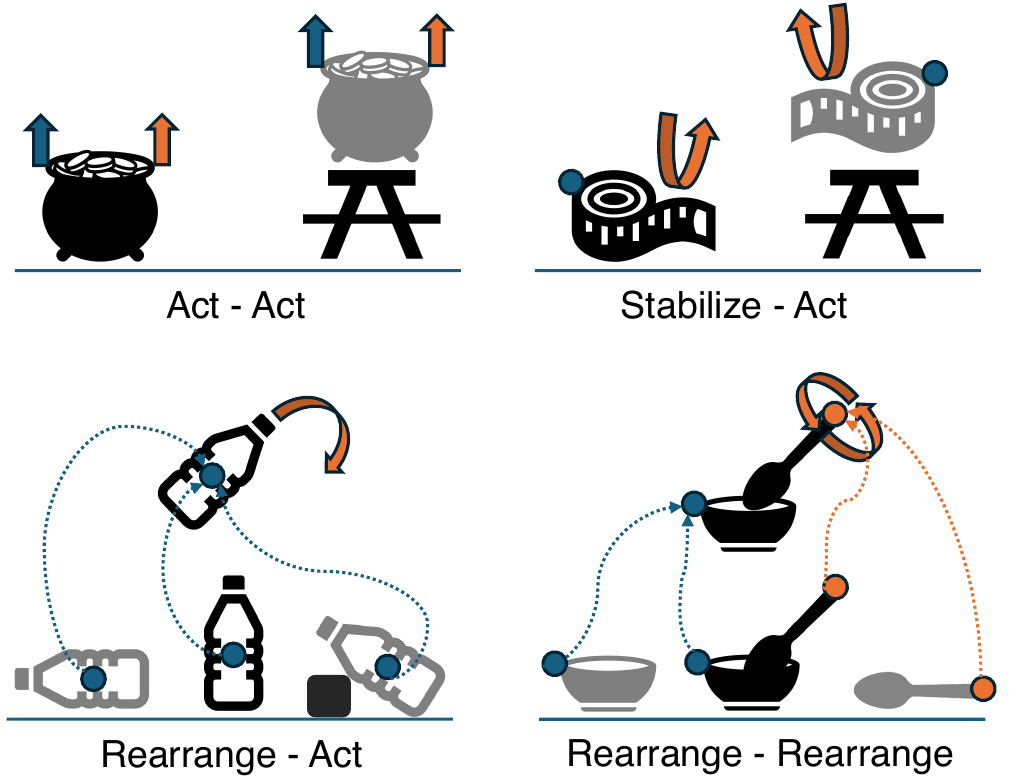}
\caption{Dual-arm Coordination Paradigm. The tasks visualized include lifting a pot, splitting tape, uncapping a bottle, and stirring a bowl. Blue denotes one arm, orange the other.}
    \label{fig:coordination}
\end{figure}

Inspired by \cite{grannen2023stabilize}, we propose a dual-arm coordination paradigm based on three fundamental manipulation primitives: \textit{act}, \textit{stabilize}, and \textit{rearrange}. The \textit{act} primitive involves interacting with an object using tailored velocity profiles; the \textit{stabilize} primitive maintains a static pose to ensure stability; and the \textit{rearrange} primitive repositions an object between poses. These primitives underpin four distinct coordination paradigms commonly observed in human manual tasks. The first paradigm, \textbf{\textit{Act-Act}}, involves both arms executing act trajectories either simultaneously or sequentially. In the \textbf{\textit{Stabilize-Act}} paradigm \cite{grannen2023stabilize}, one arm stabilizes an object while the other performs an act trajectory. The \textbf{\textit{Rearrange-Act}} paradigm is characterized by one arm repositioning an object, followed by the other executing an act trajectory. Lastly, in the \textbf{\textit{Rearrange-Rearrange}} paradigm, both arms reposition two objects, moving them back to demonstration poses for subsequent interactions. Figure \ref{fig:coordination} illustrates example everyday tasks based on this concept. For the \textit{Stabilize} and \textit{Rearrange} primitives, we record only the target EE poses, allowing motion planners to generate dynamically optimized trajectories during testing. This retains essential interaction waypoints rather than directly replicating demonstrations, enabling adaptive path generation that accounts for kinematic constraints and collision avoidance. Fundamentally, the paradigm decomposes complex multi-stage dual-arm manipulation tasks into sequences of short-horizon primitives that rely solely on \( {}^\delta \boldsymbol{T}^O_R \), \textbf{\textit{the relative pose change of the object in the robot frame from demonstration to test}}. 

Let \( \boldsymbol{T}_{R}^{B} \) be the demonstration bottleneck pose in the robot frame, and \( \boldsymbol{T}_{R}^{B'} \) be the testing bottleneck pose. Since \( \boldsymbol{T}_{O}^{B} \approx \boldsymbol{T}_{O}^{B'} \), the transformation \( {}^\delta \boldsymbol{T}^O_R \) can be implicitly estimated by:
\begin{equation}
    {^{\delta}} \boldsymbol{T}_{R}^{O} = \boldsymbol{T}_{R}^{B'} \left( \boldsymbol{T}_{R}^{B} \right)^{-1}.
\end{equation}
\indent In addition to the bottleneck parametrization depicted above, time parametrization is also required, which can be achieved using TOTG \cite{kunz2013time}, leading to a coordinated trajectory \( \boldsymbol{\tau}(B, t) \). For the \textit{Act-Act} and \textit{Stabilize-Act} tasks, \( \boldsymbol{\tau}(B, t) \) adapts in \textbf{\textit{Cartesian space}} by transforming the demonstrated EE poses—obtained via forward kinematics (FK)—using \( {^{\delta}} \boldsymbol{T}_{R}^{O} \), and subsequently solving for inverse kinematics (IK). In contrast, the \textit{Rearrange-Act} and \textit{Rearrange-Rearrange} tasks employ a hybrid approach: objects are first rearranged to their demonstrated poses through online motion planning, followed by demonstration trajectory replay in \textbf{\textit{joint space}}. This facilitates the execution of complex trajectories, such as stirring, which are challenging to imitate in Cartesian space.

In order to perform precise and coordinated tasks using these paradigms, a highly accurate estimation of \( \boldsymbol{T}_{R}^{B'} \) must be obtained at test time, which is achieved via our 3-VS.

\subsection{Three-stage Visual Servoing (3-VS)}
\label{hdmvs}
To precisely estimate \( \boldsymbol{T}_{R}^{B'} \) in novel configurations, we draw inspiration from recent advancements in \textbf{\textit{learning-based feature matching}} \cite{sarlin2020superglue, sun2021loftr, lindenberger2023lightglue, jiang2024omniglue}. These deep matchers, which utilize Transformers \cite{vaswani2017attention}, are trained to simultaneously match local features and reject outliers in image pairs. During demonstration, we obtain two bottleneck RGB-D images \( \mathcal{I}_{C_g}^B \) and \( \mathcal{I}_{C_w}^B \) from \({C_g} \) and \({C_w} \). A prediction of \( \boldsymbol{T}_{R}^{B'} \) can be obtained by matching \( \mathcal{I}_{C_g}^B \) and \( \mathcal{I}_{C_w}^B \) with their respective current images, \( \mathcal{I}_{C_g} \) and \( \mathcal{I}_{C_w} \). Let us define \( \smash{^{*}\boldsymbol{T}_{R}^{B'}} \) as the optimal bottleneck pose. \( ^{*}\boldsymbol{T}_{R}^{B'} \) can be considered as the EE pose that minimizes the discrepancy in 2D correspondences produced by the deep matchers on the task object \( O \) between \( \mathcal{I}_{C_w}^B \) and \( \mathcal{I}_{C_w} \). The challenge, therefore, is to reliably control the EE to \( \smash{^{*}\boldsymbol{T}_{R}^{B'}} \). Given the assumption of an arbitrary initial position, the estimation of \( \smash{^{*}\boldsymbol{T}_{R}^{B'}} \) requires both global and local feature information of \( O \). To address this, we decompose the problem into three stages, progressively refining the estimate. Stage One generates an initial coarse estimate by leveraging the global camera’s wide field of view. Stage Two facilitates a smooth transition of confidence from the global camera’s estimate to the wrist camera. Finally, Stage Three ensures precise alignment as the system converges, relying solely on the wrist camera. Experimental validation confirms that this three-stage approach enhances both accuracy and robustness.



\begin{figure*}[t]
\setlength{\belowcaptionskip}{-12pt}
\centering
\includegraphics[width=1\textwidth]{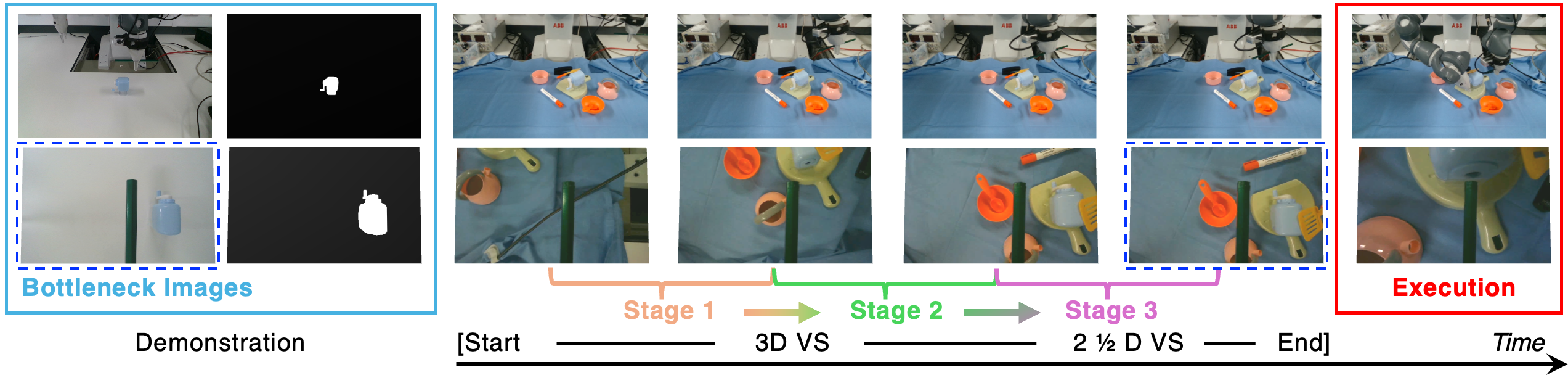}
\caption{Three-stage Visual Servoing. (1) The process begins with \textbf{\textit{3D Visual Servoing}}, using the initial open-loop global-camera bottleneck estimate for control until the wrist camera's estimates become stable. (2) Once stable, the initial global-camera estimate serves as a prior and is fused with redundant sequential wrist-camera estimates using a Kalman filter, until the overlap between the bottleneck and current wrist-camera images exceeds a threshold. (3) The control strategy then shifts to \textbf{\textit{2 1/2 D Visual Servoing}} until convergence. Finally, the coordinated trajectory is adapted and executed.}
\label{pipeline}
\end{figure*}

Figure \ref{pipeline} illustrates the pipeline with real images from both \( {C}_{g} \) and \( {C}_{w} \). In the first two stages, we use \textbf{\textit{3D visual servoing}}, which integrates multiple cameras, redundant measurements, and motion predictions into a unified system \cite{wilson1996relative}. In the third stage, we switch to \textbf{\textit{2 1/2 D visual servoing}} \cite{760345} to enhance robustness against imprecise hand-eye calibration and noise from the deep matchers and depth sensors. We explain in detail each of the stages in Section \ref{stage1}
to \ref{stage3}.




\subsection{Stage One: 3D Visual Servoing with Global Camera}
\label{stage1}
Define the mapping from the EE pose \( \boldsymbol{T}_{R}^{E} \) to its corresponding axis-angle and translation representation as \(\mathcal{M}\), such that \(\boldsymbol{x}^{E} = \mathcal{M}(\boldsymbol{T}_{R}^{E})\), and \(\boldsymbol{x}^{B'} = \mathcal{M}(\boldsymbol{T}_{R}^{B'})\). During testing, the global camera \({C_g}\) detects the task object \( O \) in novel configurations and makes an initial bottleneck estimate. This estimate is derived by estimating the relative pose change of \( O \) from demonstration to test within \({C_g}\)'s frame, denoted as \({}^\delta \smash{\hat{\boldsymbol{T}}}^O_{C_g}\). Let \( \smash{\tilde{\boldsymbol{T}}}_{R}^{{C}_{g}} \) denote the imprecise hand-eye calibration of \({C}_{g}\). \(C_g\)'s initial bottleneck estimate \( \smash{\hat{\boldsymbol{x}}_{C_g}^{B'}} \) is then given by:
\begin{equation}
\label{prior}
\hat{\boldsymbol{x}}_{C_g}^{B'} = \mathcal{M}\left( 
\tilde{\boldsymbol{T}}_{R}^{{C}_{g}} \, 
{}^\delta \hat{\boldsymbol{T}}^O_{C_g}
\left( \tilde{\boldsymbol{T}}_{R}^{{C}_{g}} \right)^{-1} 
\boldsymbol{T}_{R}^{B}\right),
\end{equation}
\indent We follow \cite{vitiello2023one} and apply partial point cloud registration (PPCR) to estimate \( {}^\delta \smash{\hat{\boldsymbol{T}}}^O_{C_g} \) using \( \mathcal{I}_{C_g}^B \) and \( \mathcal{I}_{C_g} \). Specifically, we utilize FPFH \cite{5152473}, FGR \cite{Zhou2016FastGR}, and ICP (justified in Section \ref{deepmatchexperiments}). This process necessitates a live segmentation of \( O \) in \( \mathcal{I}_{C_g} \). We employ GroundedSAM \cite{ren2024grounded} for open-vocabulary instance segmentation, which negates the need for object-specific detectors. At stage one only, PPCR is preferred over deep matching due to \( O \)'s small size relative to \( \mathcal{I}_{C_g} \). Sparse matchers, such as \cite{lindenberger2023lightglue}, often yield insufficient matches for small segmented objects, while dense matchers like \cite{edstedt2024roma} are too resource-intensive and slow for our GPU. In this context, inference time is also critical: if \( O \) is displaced during testing, \(C_g\) must quickly respond and reinitialize the 3-VS controller to the first stage if necessary, as \(C_w\) may lose sight of \( O \).


In the first stage, a decoupled proportional control law based on error \( \boldsymbol{e} = {\boldsymbol{x}}^E - \smash{\hat{\boldsymbol{x}}_{C_g}^{B'}} \) is used for \textbf{\textit{3D Visual Servoing}}. The objective is to move the EE closer to \( O \) to stabilize \(C_w\)'s bottleneck estimates, which use deep matching due to the richer, high-resolution features available for matching between \( \mathcal{I}_{C_w}^B \) and \( \mathcal{I}_{C_w} \). Stabilization occurs when the number of confident matches between \( \mathcal{I}_{C_w}^B \) and \( \mathcal{I}_{C_w} \) exceeds a threshold, and the moving variance of \( C_w \)'s bottleneck estimates falls below another experimentally determined threshold.

\subsection{Stage Two: 3D Visual Servoing with Both Cameras}
\label{stage2}

In the second stage, \(C_w\) contributes to more accurate bottleneck estimates by reconstructing its relative displacement in \(O\)'s frame from the bottleneck to the current instance, \( {}^\delta \smash{\hat{\boldsymbol{T}}}^{C_w}_{O} \), using \( \mathcal{I}_{C_w}^B \) and \( \mathcal{I}_{C_w} \). Unlike \(C_g\), \(C_w\) remains in continuous motion, causing \(O\) to be non-stationary in \( \mathcal{I}_{C_w} \). Let \( \smash{\tilde{\boldsymbol{T}}}_{E}^{{C}_{w}} \) represent \(C_w\)'s imprecise hand-eye calibration. At time step \( k \), \(C_w\)'s bottleneck estimate \( \hat{\boldsymbol{x}}_{C_w, k}^{B'} \) is given by:
\begin{equation}
\label{measurement}
 \hat{\boldsymbol{x}}_{C_w, k}^{B'} = \mathcal{M}\left( \boldsymbol{T}_{R}^{E} \, \tilde{\boldsymbol{T}}_{E}^{{C}_{w}} \,
{}^\delta \hat{\boldsymbol{T}}^{C_w, k}_{O}
\left( \tilde{\boldsymbol{T}}_{E}^{{C}_{w}} \right)^{-1}\right),
\end{equation}
\indent The estimation of \( {}^\delta \smash{\hat{\boldsymbol{T}}}^{C_w}_{O} \) is performed using sparse deep matching with SIFTGPU \cite{lowe1999object, Wu2007SiftGPUA} and LightGlue (LG) \cite{lindenberger2023lightglue} (justified in Section \ref{deepmatchexperiments}), using \( \mathcal{I}_{C_w}^B \) and \( \mathcal{I}_{C_w} \). Since \( \mathcal{I}_{C_w}^B \) also contains background that may be matched incorrectly, we apply \textbf{\textit{unidirectional filtering}} to discard matches outside the segmentation mask of \( \mathcal{I}_{C_w}^B \). The correspondences are then lifted to 3D by aligning with depth images, and \( {}^\delta \smash{\hat{\boldsymbol{T}}}^{C_w}_{O} \) is estimated using the weighted Kabsch algorithm \cite{Kabsch:a15629} using weights from LG normalized, and with RANSAC applied iteratively to further improve robustness to incorrect matches.


Given the redundancy in measurements and the non-linearity, we employ an \textbf{\textit{Unscented Kalman Filter}} \cite{882463} to enhance the reliability and accuracy of the bottleneck estimates. The prior \(\hat{\boldsymbol{x}}_{C_g}^{B'}\) is fused with measurements \(\hat{\boldsymbol{x}}_{C_w, k}^{B'}\) to produce the posterior estimate \(\bar{\boldsymbol{x}}^{B'}_{k \mid k}\). The control law is the same as the first stage, but based on error \(\boldsymbol{e}_k = ({\boldsymbol{x}}^{E} - \bar{\boldsymbol{x}}^{B'}_{k \mid k})\).

We transition to the third stage once \( O \) exhibits sufficient overlap between \( \mathcal{I}_{C_w}^B \) and \( \mathcal{I}_{C_w} \), indicated by a consistently small posterior \( {}^\delta \smash{\bar{\boldsymbol{T}}}^{C_w}_{O} \) over a short period. This transition mitigates inaccuracies from \( \smash{\tilde{\boldsymbol{T}}}_{E}^{{C}_{w}} \), noisy depth images, and mismatches, while enhancing robustness against outliers.

\subsection{Stage Three: 2 1/2 D Visual Servoing with Wrist Camera}
\label{stage3}
In the third stage , only \(C_w\) is used, as by this point the EE should be close to \( \smash{^{*}\boldsymbol{T}_{R}^{B'}} \). The goal is to achieve the most accurate and robust estimation possible. This is accomplished through \textbf{\textit{2 1/2 D Visual Servoing}}, which leverages matched features to control the robot’s motion at the image level, along with reconstructed information \cite{760345}. A homography matrix for a virtual plane attached to \(O\) is estimated and regulated towards the identity matrix \cite{chaumette2007visual}. We employ a variation from \cite{999646} that improves tolerance to calibration errors. Let \(\theta \mathbf{u}\) denote the rotational component of \( \mathcal{M}({}^\delta \smash{\hat{\boldsymbol{T}}}^{C_w}_{O}) \). The task function \(\mathbf{e}\) is then given by:
\begin{equation}
\label{2.5Dtaskfunction}
    \quad \mathbf{e} = \bigm[\rho \mathbf{m} - \mathbf{m}^{*} \quad \rho - 1 \quad \theta\mathbf{u} \bigm]^{T},
\end{equation}
\indent Where \( \mathbf{m}^{*} = \begin{bmatrix} \mathbf{x}^{*} & \mathbf{y}^{*} & 1 \end{bmatrix}^{T} \) is the vector containing the homogeneous coordinates of a fixed reference point obtained from \( \mathcal{I}_{C_w}^B \), and \(\rho =  Z / Z^{*} \) represents the ratio of the current point's depth to \(\mathbf{m}^{*}\)'s depth. At this stage, we use SuperPoint (SP) \cite{detone2018superpoint} along with LG \cite{lindenberger2023lightglue} for feature matching (justified in Section \ref{deepmatchexperiments}). Following \cite{lindenberger2023lightglue}, we estimate the homography matrix \(\mathbf{H}\) using MAGSAC++ \cite{barath2020magsac++}, and select \(\mathbf{m}^{*}\) from the inlier set with the smallest reprojection error, as this point is most likely to lie on the virtual plane. We then compute \(\mathbf{m}\) and \(\rho\) with the aligned depth images as follows:
\begin{equation}
\label{homography}
\rho \, \mathbf{m} = \mathbf{H} \, \mathbf{m}^{*},
\end{equation}
\indent The reconstruction of \(\theta\mathbf{u}\) is achieved by decomposing $\mathbf{H}$:
\begin{equation}
\label{H_decompose}
\mathbf{H} = \boldsymbol{R} + \frac{\boldsymbol{t}}{d^{*}} \mathbf{n}^{*\top},
\end{equation}
\indent where \(\mathbf{t}\) is the translation vector (up to a scale), \(\mathbf{n}^{*}\) is the normal vector, and \(d^{*}\) is the distance \cite{Malis200021D}. The decomposition yields two admissible solutions after Cheirality test \cite{benhimane2007homography}. We select the correct solution based on temporal consistency.

When the error \(\mathbf{e}\) falls below a threshold, we apply \textbf{\textit{bi-directional filtering}}, assuming that the segmentation mask of \( \mathcal{I}_{C_w}^B \) also applies to \( \mathcal{I}_{C_w}\). This approach further reduces noise from distractors and partial occlusions, enhancing robustness.


\section{Experiments}



\subsection{Deep Image Matching for Visual Servoing}
\label{deepmatchexperiments}
3-VS consists of three stage, each requiring the selection of an optimal matching method. This section explains how we identified the most suitable method for each stage.

\textbf{\textit{Stage One}} is an open-loop stage that uses only the global camera's images. Dense matchers such as ROMA \cite{edstedt2024roma} and DKM+GIM \cite{edstedt2023dkm, shen2024gim} perform well in this context but are too resource-intensive for our GPU. As a result, we use PPCR for efficiency. When dealing with small partial object point clouds, we found that a traditional pipeline using \textbf{\textit{FGR}} \cite{Zhou2016FastGR} \textbf{\textit{+}} \textbf{\textit{ICP}} outperforms learning-based methods such as \cite{wang2019deep, aoki2019pointnetlk, vitiello2023one}, which make unrealistic assumptions, as also noted in \cite{Lee2021DeepPRODP}.


\textbf{\textit{Stage Two}} is a closed-loop state that now integrates estimates from the wrist camera using sparse matching. While these matchers have been benchmarked \cite{jiang2024omniglue, mast3r_arxiv24}, our focus is on the manipulation domain with object-centric images, and also assessing the matching uncertainty across varying viewpoints for reliable control. To this end, we collected a real-world dataset using an eye-in-hand RealSense D405 camera (480 × 848 resolution) to capture images from 26 EE poses, including bottleneck poses and additional poses with translations ranging from 3 to 15 cm (in 3 cm increments) and rotations from 15° to 75° (in 15° increments), resulting in 4-DoF variations. For each pose, 50 RGB-D images were captured for 12 objects, totaling 15.6k images. \( {}^\delta \smash{\hat{\boldsymbol{T}}}^{C_w}_{O}\) was estimated using the method described in Section \ref{stage2}, without RANSAC or weights. Ground truth was computed using Equation \ref{measurement}, and the results were assessed using the mean isotropic rotation and translation error.

\begin{table}
\centering
\setlength{\belowcaptionskip}{-15pt}
\renewcommand{\arraystretch}{1.1} 
\setlength{\tabcolsep}{1.9pt} 
\begin{tabular}{lcccccccccc}
\toprule
\multirow{2}{*}{Method} & \multicolumn{4}{c}{\textbf{\textls[50]{Translation}} (cm) \hspace{0.2em} $\downarrow$} & \multicolumn{4}{c}{\textbf{\textls[50]{Rotation}} (\textdegree) \hspace{0.2em} $\downarrow$ } & \multirow{2}{*}{FPS} \\
\cmidrule(lr){2-5} \cmidrule(lr){6-9}
& CV & ME & CRE & SSE & CV & ME & CRE & SSE \\
\midrule
DINO\cite{caron2021emerging} & \multirow{2}{*}{0.82} & \multirow{2}{*}{21.8} & \multirow{2}{*}{10$\pm$9.4} & \multirow{2}{*}{1.4$\pm$1.0} & \multirow{2}{*}{0.53} & \multirow{2}{*}{47.5} & \multirow{2}{*}{24$\pm$14} & \multirow{2}{*}{3.4$\pm$2.4} & \multirow{2}{*}{0.5}\\
+MNN \\ 
XFeat*\cite{potje2024xfeat} & \multirow{2}{*}{0.58} & \multirow{2}{*}{39.0} & \multirow{2}{*}{14$\pm$10} & \multirow{2}{*}{\underline{0.7$\pm$0.3}} & \multirow{2}{*}{\underline{0.48}} & \multirow{2}{*}{85.3} & \multirow{2}{*}{38$\pm$24} & \multirow{2}{*}{\underline{1.6$\pm$0.8}}  & \multirow{2}{*}{\textbf{27}} \\
+MR \\ 
SP\cite{detone2018superpoint} & \multirow{2}{*}{\underline{0.50}} & \multirow{2}{*}{28.5} & \multirow{2}{*}{\underline{6.8$\pm$5.3}} & \multirow{2}{*}{0.9$\pm$0.5} & \multirow{2}{*}{\textbf{0.46}} & \multirow{2}{*}{51.4} & \multirow{2}{*}{\underline{13$\pm$9.4}} & \multirow{2}{*}{2.0$\pm$1.1} & \multirow{2}{*}{\underline{23}} \\
+LG\cite{lindenberger2023lightglue} \\
SP+LG & \multirow{2}{*}{0.63} & \multirow{2}{*}{\textbf{15.6}} & \multirow{2}{*}{8.7$\pm$10} & \multirow{2}{*}{\textbf{0.5$\pmb{\pm}$0.3}} & \multirow{2}{*}{0.66} & \multirow{2}{*}{\underline{31.3}} & \multirow{2}{*}{19$\pm$19} & \multirow{2}{*}{\underline{1.6$\pm$0.8}} & \multirow{2}{*}{17} \\
+GIM\cite{shen2024gim}  \\
SIFT\cite{Wu2007SiftGPUA} & \multirow{2}{*}{\textbf{0.47}} & \multirow{2}{*}{\underline{16.8}} & \multirow{2}{*}{\textbf{4.3$\pmb{\pm}$1.5}} & \multirow{2}{*}{\textbf{0.5$\pmb{\pm}$0.3}} & \multirow{2}{*}{0.64} & \multirow{2}{*}{\textbf{22.7}} & \multirow{2}{*}{\textbf{6.5$\pmb{\pm}$4.0}} & \multirow{2}{*}{\textbf{1.3$\pmb{\pm}$0.8}} & \multirow{2}{*}{22} \\
+LG \\
Mast- & \multirow{2}{*}{0.52} & \multirow{2}{*}{44.7} & \multirow{2}{*}{35$\pm$24} & \multirow{2}{*}{4.1$\pm$9.7} & \multirow{2}{*}{0.49} & \multirow{2}{*}{64.5} & \multirow{2}{*}{53$\pm$35} & \multirow{2}{*}{9.6$\pm$17} & \multirow{2}{*}{1.4} \\ 
3R\cite{dust3r_cvpr24} \\
\bottomrule
\end{tabular}
\caption{Comparison of recent sparse feature matching methods for relative pose estimation in a tabletop, object-centric setting. Metrics include Coefficient of Variation (CV), Mean Error (ME), Close Region Error (CRE), and Static State Error (SSE). Best results highlighted, second-best underlined.}
\label{tab:correspondence_comparison}
\end{table}

\begin{figure}[b]
    \vspace{-10pt}
    \centering
    \includegraphics[width=0.48\textwidth]{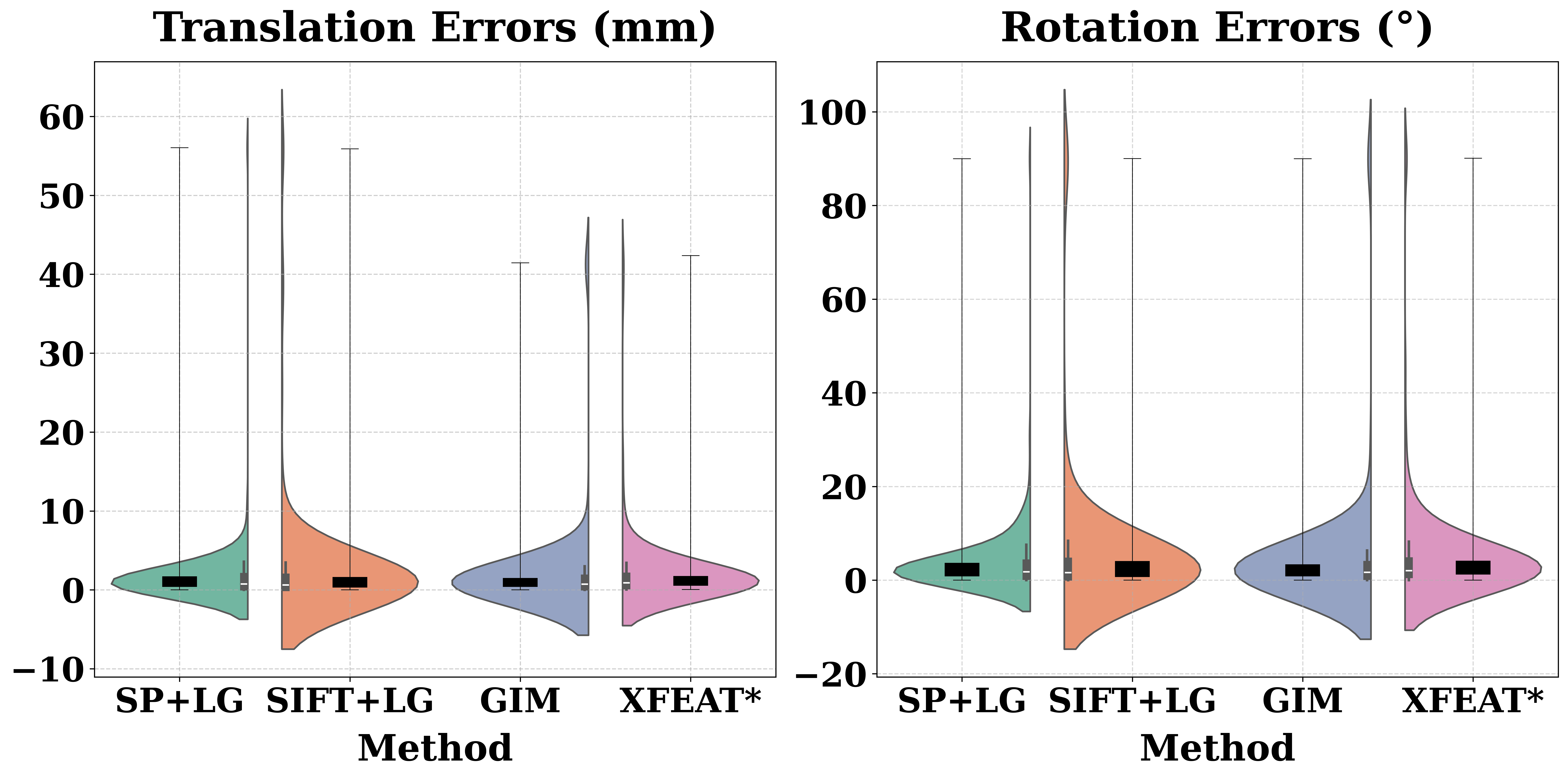}  
    \caption{Comparison of the best 4 methods from Table \ref{tab:correspondence_comparison} near convergence with strong illumination variations.}
    \label{fig:illumination}
\end{figure}

\begin{table*}[t!]
    \centering
    \setlength{\tabcolsep}{1.8pt}
    \caption{Real-world experiment results, displaying the percentage success rate over 10 trials for each cell.}
    \begin{tabular}{lcccccccccccccccccccccc}
        \toprule
        Method & \multicolumn{6}{c}{\textbf{\textls[50]{Act - Act}}} & \multicolumn{6}{c}{\textbf{\textls[50]{Stabilize - Act}}} & \multicolumn{6}{c}{\textbf{\textls[50]{Rearrange - Act}}} & \\
        \cmidrule(lr){2-7} \cmidrule(lr){8-13} \cmidrule(lr){14-19}
        & \multicolumn{3}{c}{\textbf{Lift Rack}} & \multicolumn{3}{c}{\textbf{Place in Cooker}} & \multicolumn{3}{c}{\textbf{Uncover Scissor}} & \multicolumn{3}{c}{\textbf{Insert Pencil}} & \multicolumn{3}{c}{\textbf{Split Blocks}} & \multicolumn{3}{c}{\textbf{Hang Pan}} & Avg \\
        \cmidrule(lr){2-4} \cmidrule(lr){5-7} \cmidrule(lr){8-10} \cmidrule(lr){11-13} \cmidrule(lr){14-16} \cmidrule(lr){17-19} 
        & 4DoF & 4DoF$^{\text{+}}$ & 6DoF$^{\text{+}}$ & 4DoF & 4DoF$^{\text{+}}$ & 6DoF$^{\text{+}}$ & 4DoF & 4DoF$^{\text{+}}$ & 6DoF$^{\text{+}}$ & 4DoF & 4DoF$^{\text{+}}$ & 6DoF$^{\text{+}}$ & 4DoF & 4DoF$^{\text{+}}$ & 6DoF$^{\text{+}}$ & 4DoF & 4DoF$^{\text{+}}$ & 6DoF$^{\text{+}}$ \\
        \midrule
        FGR+ICP\cite{vitiello2023one} & 40 & 30 & 30 & 0 & 0 & 0 & 20 & 30 & 0 & 0 & 0 & 0 & 20 & 0 & 10 & 0 & 0 & 0 & 10 \\
        RoboTAP\cite{vecerik2024robotap}  & 40 & 20 & - & 20 & 10 & - & 70 & 50 & - & 0 & 0 & - & 40 & 40 & - & 0 & 0 & - & 24.2 \\
        DINOBot\cite{dipalo2024dinobot}  & 70 & 50 & 40 & 40 & 30 & 0 & 20 & 0 & 0 & 0 & 0 & 0 & 30 & 20 & 30 & 0 & 0 & 0 & 18.3 \\
        \rowcolor{LightCyan}
        Ours & 100 & 90 & 70 & 100 & 100 & 60 & 100 & 90 & 60 & 80 & 50 & 20 & 90 & 80 & 60 & 90 & 90 & 50 & 77.2 \\
        \bottomrule
    \end{tabular}
    \label{tab:tasks_comparison}
    \vspace{-5pt}
\end{table*}

\begin{figure*}[t!] 
\setlength{\belowcaptionskip}{-15pt}
\centering
\begin{minipage}{\textwidth}
    \centering
    \includegraphics[width=0.75\textwidth]{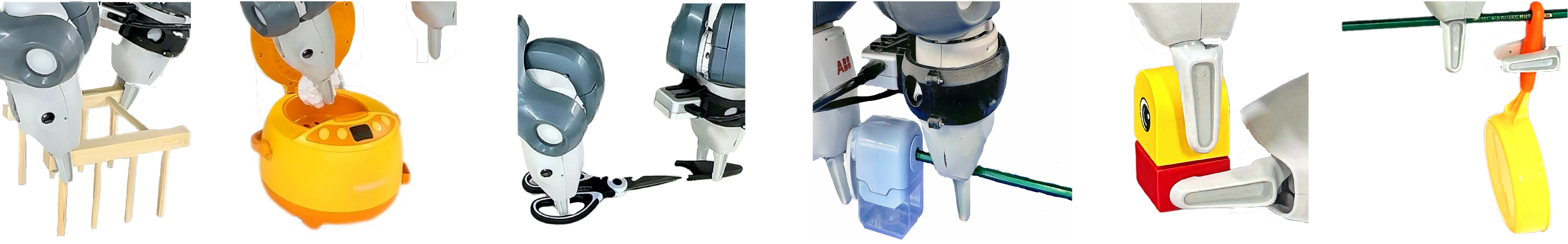}
    \captionof{figure}{The 6 dual-arm tasks listed in Table \ref{tab:correspondence_comparison}.}
    \label{realworldtasks}
\end{minipage}
\end{figure*}

Table \ref{tab:correspondence_comparison} presents the results. The Coefficient of Variation (CV) measures consistency across the 50 images at each pose, with an ideal CV of 0 indicating perfect matchers and noiseless sensors. Lower CV values suggest more stable estimates but do not necessarily imply higher accuracy. The Mean Error (ME), averaged over 26 viewpoints, evaluates robustness to varying viewpoints, with lower ME and CV values being preferable. This is particularly important when starting from poses significantly different from the bottleneck pose, such as translations of 15 cm and rotations of 75°, where the object is barely visible in the wrist camera. However, the use of a global camera for prior estimation reduces the need for extreme robustness. The Close Region Error (CRE) measures accuracy in confined spaces, specifically for translations of 3-6 cm and rotations of 15-45° around the bottleneck pose. The Static State Error (SSE) assesses accuracy at the exact bottleneck pose. FPS is measured using a RTX 3070, with all methods optimized to avoid repetitive feature extraction and description from the bottleneck image.

We observe that learned matchers generally outperform classical ones, which often struggle with scarce matches and extreme outliers. However, learned features often underperform compared to SIFT, making \textbf{\textit{SIFT+LG}} the best. \textbf{\textit{This disparity arises because state-of-the-art feature matchers are sensitive to rotations, having been predominantly trained on upright images with an implicit assumption of gravity alignment}}. We also experimented with Dust3R \cite{dust3r_cvpr24}. However, with only two views, the model frequently outputted identity.



\textbf{\textit{Stage Three}} is a closed-loop stage that uses only the images from the wrist camera. Even the best method from Table \ref{tab:correspondence_comparison} shows SSE translation uncertainties of 0.3 cm due to matcher and sensor noise. Lighting conditions, which were not previously addressed, also affect performance. To account for this, we collected an additional dataset near the bottleneck pose under varying illumination conditions. The results, shown in Figure \ref{fig:illumination}, indicate that \textbf{\textit{SP+LG}} achieved the sharpest distribution and provided more matches in this experiment, now with smaller rotations. However, all methods exhibit extreme outliers, likely due to numerical instabilities and degenerate cases, such as co-planar points. This further emphasizes the importance of stage three, where robust homography-based control is employed, with part of the task function directly expressed on the image planes.

\subsection{Real-world Robotic Evaluation}
We now compare our overall framework, ODIL, to state-of-the-art one-shot imitation learning methods that also utilize VS, RoboTAP \cite{vecerik2024robotap} and DINOBot \cite{dipalo2024dinobot}. The evaluation encompasses challenging scenarios including background variations, distractions, partial occlusions, and 6-DoF object pose changes. We did not include behavior cloning methods in our comparison, as they have been shown to perform poorly with a single demonstration \cite{dipalo2024dinobot, valassakis2022dome}. FGR+ICP was also tested following the previous experiment as an open-loop baseline, using only the initial estimate from the global camera for VS. \cite{vitiello2023one} did not perform well in our setup and would require training a custom dataset tailored to our global-camera extrinsic. Instead, we opted for FGR+ICP, which \cite{vitiello2023one} reported to yield similar performance.

\label{realwordexp}
\textbf{System setup} includes an ABB YuMi robot with a position controller operating at 10 Hz. The Cartesian controller is implemented using TRAC-IK \cite{7363472}, with collision and singularity avoidance managed by MoveIt \cite{coleman2014moveit}. We utilize a wrist-mounted RealSense D405 camera (480 x 848), and a global D415 camera (720 x 1280) mounted on the table. 

\textbf{Experiments} were done with 6 precise and coordinated dual-arm tasks (Figure \ref{realworldtasks}). Each task was evaluated under 3 conditions: first, 4-DoF object pose changes with a uniform background; second, 4-DoF$^{\text{+}}$ which also includes background changes, distractor objects, and partial occlusion; and third, 6-DoF$^{\text{+}}$ with 6-DoF object pose changes plus background changes, distractor objects, and partial occlusion. We conducted 10 trials per task for each condition. In the 4-DoF experiments, we used 4-DoF EE control, with the object’s rotation around the z-axis up to ±60° relative to the demonstration orientation. In the 6-DoF experiments, we used 6-DoF EE control, and the object’s rotations around the x and y axes are up to ±30°. Since RoboTAP and DINOBot rely on a single wrist camera, we manually initialize the EE to approximately 10 cm above the bottleneck to ensure the object's visibility. As our method does not assume that objects are initially visible in the wrist camera’s view, we re-run the controller for our method if the object is not visible after stage one, for a fair comparison to the baselines. The controller parameters were tuned once to accommodate the illumination conditions, and applied across all tasks.

\textbf{Results} are shown in Table \ref{tab:tasks_comparison}. Our method consistently outperformed the baselines across all settings and tasks. Although there was a minimal drop in performance when transitioning from 4-DoF to 4-DoF$^{\text{+}}$, the pencil insertion task was an exception. In this task, repeated failures occurred when the pencil held by the gripper overlapped with partial occlusions on the sharpener, leading to significant obstruction and causing the controller to become idle. However, we observed that the controller occasionally recovered from these situations over time. Additionally, we found that distractors, particularly objects with similar appearance to the task object, caused all controllers to exhibit increased jerkiness.

A notable performance drop was observed in the 6-DoF$^{\text{+}}$ experiments. In these cases, stage two often reached undesired EE poses due to poor prior estimation from the global camera, and frequently failed to recover due to minimal overlap between the bottleneck and current wrist camera images. This highlighted the challenges of controlling in \textit{SE(3)} without a complete 3D object representation. However, for trials that did progress to stage three, the success rate remained high. Nevertheless, we observed a decline in precision compared to the 4-DoF$^{\text{+}}$ settings, which can be attributed to the additional degrees of freedom that needed to be precisely estimated. Among all the methods, FGR+ICP had the lowest success rate due to being limited to open-loop VS. While DINOBot was less precise than RoboTAP, we observed that it exhibited greater robustness to object rotations. 



Videos from all tasks and settings, together with failure modes and further experiments, are available on our website: \begingroup
\hypersetup{urlcolor=blue}
\url{https://www.robot-learning.uk/one-shot-dual-arm}
\endgroup.



\section{Conclusion}
We have introduced ODIL, a novel method for dual-arm robots to learn precise and coordinated tasks from a single demonstration, leveraging a dual-arm coordination paradigm and a three-stage visual servoing controller. We benchmarked recent deep feature matching methods for visual servoing and demonstrated that combining these methods with traditional analytic techniques achieves the best performance. Real-world experiments showed that ODIL outperforms other state-of-the-art one-shot imitation learning methods and remains robust under challenging scene changes. Opportunities for future work include extending to multi-stage tasks, introducing failure recovery and closed-loop object interaction, and generalising learned behaviours to novel objects.


\clearpage


{\small
\printbibliography
}

\end{document}